\documentclass{svproc}
\usepackage{url}
\usepackage{amsfonts}       
\usepackage{amssymb}
\usepackage{amsmath}\usepackage{graphicx}
\usepackage{multirow,tabularx}
\newcolumntype{C}{>{\centering\arraybackslash}X}

\newcommand\Tstrut{\rule{0pt}{2.5ex}}       
\newcommand\Bstrut{\rule[-1.1ex]{0pt}{0pt}} 
\newcommand{\TBstrut}{\Tstrut\Bstrut} 
\usepackage{makecell}

\usepackage[parfill]{parskip}

\begin{document}
\mainmatter              
\title{Deep Reinforcement Learning for Control of
Probabilistic Boolean Networks\thanks{This research was partly funded by EIT Digital IVZW, under the Real-Time Flow project, activity 18387-SGA2018, and partly by the EPSRC project AGELink (EP/R511791/1). We would also like to thank Vytenis Sliogeris for implementing the PBN inference pipeline from gene-expression data of the metastatic-melanoma.}}
\titlerunning{ }
%
%
\author{Georgios Papagiannis \and Sotiris Moschoyiannis}
%
\authorrunning{ }
\institute{University of Surrey, UK\\
\email{\{g.papagiannis, s.moschoyiannis\}@surrey.ac.uk}}

\maketitle              

\begin{abstract}
Probabilistic Boolean Networks (PBNs) were introduced as a computational model for the study of complex dynamical systems, such as Gene Regulatory Networks (GRNs). {\it Controllability} in this context is the process of making strategic interventions to the state of a network in order to drive it towards some other state that exhibits favourable biological properties. In this paper we study the ability of a Double Deep Q-Network with Prioritized Experience Replay in learning control strategies within a finite number of time steps that drive a PBN towards a target state, typically an attractor. The control method is model-free and does not require knowledge of the network's underlying dynamics, making it suitable for applications where inference of such dynamics is intractable. We present extensive experiment results on two synthetic PBNs and the PBN model constructed directly from gene-expression data of a study on metastatic-melanoma.
\end{abstract}
\section{Introduction}
The computational model of Boolean Networks (BNs) was originally introduced to model gene interactions in gene regulatory networks (GRNs). Probabilistic Boolean Networks (PBNs) \cite{schmulevich-pbn}, extended the framework of BNs to account for the uncertainty of gene interaction inherent to the model selection and data collection process. Both PBNs and BNs have been extensively used to model well-known regulatory networks, such as that of the {\it metastatic-melanoma} \cite{bittner-2000-melanoma} and {\it drosophila melanogaster} \cite{albert-2003-topology}.

Genes in biological systems have been shown to exhibit sudden emergence of ordered collective behavior \cite{huang-2000-collective-behaviour} which is manifested in PBNs as irreducible sets of states and absorbed states, also known as attractors \cite{shmulevich-2002-gene-perturbation}. Correspondence to such attractors has been observed in biological cell functions such as growth or quiescence \cite{huang-2000-collective-behaviour,huang-1999-growth}. Controllability in complex dynamical systems refers to the ability to guide a system’s behaviour towards a desired state \cite{liu-2011-controllability,cornelius-2013-realistic}. A fundamental property of networks is that perturbations to one node can affect other nodes, potentially causing the entire system to change behaviour. In the context of GRNs, controllability manifests as the process of discovering strategies to cause such perturbations by means of targeted interventions to the state of a cell (gene expression) aiming to drive it from its current state to a target state (typically, an attractor) that exhibits desirable biological properties.

In this paper, we explore this controllability problem by trying to answer the following question: "\emph{What is the series of required interventions to drive a PBN from any state towards a target attractor, in a specified intervention horizon, while being the least intrusive to the network.}". That is, we aim to find a control strategy that successfully drives the network to the target attractor by allowing at most \emph{one} gene perturbation in each state evolution of the PBN - given a maximum number of allowed interventions, often referred to as the treatment horizon \cite{datta-2003-control}. To address our control problem we apply Double Deep Q-Learning (DDQL) \cite{ddql} with Prioritized Experience Replay (PER) \cite{schaul-2015-PER}, a model-free reinforcement learning (RL) method proposed by \cite{ddql}. The approach develops the control strategy agnostic to the dynamics of the network, making it suitable for applications where inference of state transition probabilities is intractable or only the target attractor is known.

Our contributions in this work are outlined as follows: (1) We extend the framework of \emph{learning} to control as previously studied in the application of BNs \cite{papagiannis-2019-drl-rbn,karlsen-2018-evolution} to the framework of PBNs; (2) We apply DDQ-Learning with PER \cite{ddql,schaul-2015-PER} to address the problem of learning how to drive a network to a target attractor in the context of PBNs; (3) We demonstrate successful control strategies developed after training a Double Deep Q-Network (DDQN) with PER through extensive experiments on two highly stochastic synthetic PBNs with 10 and 20 nodes and a PBN model of metastatic-melanoma inferred directly from gene expression data.

The rest of the paper is organised as follows. Section 2 briefly reviews related work. Section 3 provides the necessary background. Section 4 discusses the control problem and link with DDQN and PER. Finally, Section 5 demonstrates our experiment results after applying the DDQN with PER to the problem of PBN control. Concluding remarks are included in Section 6.

\section{Related Work}

Previous work has explored the problem of controlling PBNs and BNs from multiple directions  \cite{datta-2007-intervention-survey,par-2006-pbn7,liu-2011-controllability,liu-2012-discrete-Markov,pita-canalization,choo-2018-control-nodes,papagiannis-2019-drl-rbn,toyoda-2019-time-feedback,wu-2019-policy-iteration-stochastic}. Slight variations on the type of control have allowed for methods to be developed that determine control strategies by either allowing interventions on all or some pre-specified nodes. Further, motivated by the biological properties found in various target states, different approaches perturb the states of a PBN in order to either drive it to some state or attractor in finite steps, or change the PBN's long-run behaviour by affecting its steady-state distribution under targeted interventions. Multiple control methods have been studied for their ability to control PBNs under different control type frameworks. 

The dynamical behaviour of a PBN can be studied under Markov Chain theory and many methods have been developed that take advantage of the Markov properties inferred directly from knowledge of a network's transition dynamics. \cite{datta-2003-control} uses dynamic programming, \cite{liu-2012-discrete-Markov} suggests a probability function for comparing the underlying MDPs, \cite{wu-2019-policy-iteration-stochastic} develops a policy iteration-type algorithm while \cite{shmulevich-2002-gene-perturbation} introduced the concept of {\it mean first passage time} to determine the genes that would probabilistically minimize the time steps required for desired state transitions to occur. However such methods suffer from the expensive step of inferring or utilizing such dynamics which can be intractable in large state spaces \cite{bellman-DPbook} and hence impractical.

Reinforcement learning methods have been studied on the problem of controlling PBNs and its variants. Such work includes fitted Q-Iteration \cite{sootla-2013-fittedQ}, Batch Reinforcement Learning (BRL) \cite{sirin-2013-BatchRL}, the use of Q-Learning \cite{faryabi-2007-RL-GRNs,amol2020} and rule-based reinforcement learning (XCS) trained with a variant of Q-Learning \cite{karlsen-2018-evolution}. However, little work has been done to leverage the advantages of neural-network based machine learning approaches that are efficiently scalable and can extract useful state representations. Hence, in this work, we study the \textit{learning-for-control} ability of a DDQ-Network \cite{ddql} with PER \cite{schaul-2015-PER}, a machine learning approach that leverages the benefits of neural network optimization algorithms. The control strategies are learned in a model-free manner, bypassing the scalability issues faced by model-based methods that depend on knowledge or inference of the network's transition dynamics.

\section{Background}
\subsection{Probabilistic Boolean Networks }We consider PBNs \cite{schmulevich-pbn} comprised of $n$ nodes, representing the $n$ genes found in gene regulatory networks. Each node takes values $\{0,1\}$, denoting whether a gene is unexpressed or expressed respectively. The gene expression of a PBN can be represented as a vector of boolean values $\mathbf{e}=\{g_1, g_2, ..., g_n\}$, where $g_i\in\{0,1\}$ denotes the expression level of node $n_i$. The number of possible gene expressions for a PBN is $2^n$. Each $n_i$ is comprised of a set of boolean functions $\mathbf{F}_i= \{f^{(i)}_1, f^{(i)}_2, ..., f^{(i)}_l\}$, where $f^{(i)}_k:\{0,1\}^T\rightarrow\{0,1\}$, where $l$ is the number of boolean functions and $T$ the number of input nodes to each $n_i$. The interaction of genes in PBNs is modelled by series of state evolutions in discrete time steps, where at each time step the expression level of a node is determined by its corresponding input genes and selected boolean function. At time step $t$ every node $n_i$ is assigned a boolean function $f^{(i)}_k\in\mathbf{F}_i$ with some probability $p^{(i)}_k$. Hence, the probability of a set of boolean functions being selected corresponds to: $p^{(1)}_x\cdot p^{(2)}_y\dots\cdot p^{(n)}_z$, where $x,y,...,z\in\{k:0<k\leq l\}$. Then gene expression $g_i^{'}$ at time step $t+1$ for node $n_i$ is determined by $g_i^{'}=f^{(i)}_k(g_1, g_2, ..., g_T)$. This process is applied to all $n$ nodes determining the expression level of the PBN at the next time step. Each realization of boolean functions leads to a specific next state and the fact that different realizations occur under different probabilities results in stochastic state evolution of the PBN.

\subsection{Reinforcement Learning } We consider a Markov Decision Process (MDP) which is defined as a tuple \(\{\mathcal{S}, \mathcal{A}, \mathcal{P}, r, \gamma\}\), where \(\mathcal{S}\) is a set of states, \(\mathcal{A}\) is a set of possible actions an agent can take on the environment, \(\mathcal{P}:\mathcal{S}\times\mathcal{A}\times\mathcal{S}\rightarrow[0,1]\) is a transition probability matrix, \(r:\mathcal{S}\times\mathcal{A}\rightarrow\mathbb{R}\) is a reward function and \(\gamma\in(0, 1)\) is a discount factor. In the context of PBNs, the set of states \(\mathcal{S}\) corresponds to the set of possible gene expressions $\{0,1\}^n$ comprising $2^n$ states. The set of actions \(\mathcal{A}\) corresponds to the set of allowed gene interventions. \(\mathcal{P}\) is determined by the probability of realizing each of the PBN's boolean functions combinations under the influence of some perturbation from the set of actions \(\mathcal{A}\). 

Further, we consider  \(Q^\pi(s_t, a_t)= \mathbb{E}_{s_{t+1}, a_{t+1}, ...}[\sum_{m=0}^{H-t-1} \gamma^tr(s_{t+m}, a_{t+m})]\) the state-action value function, where \(s_t\in\mathcal{S}\) is a state observed by the agent at time step \(t\) and $H$ the intervention horizon. The agent's behavior during training is defined by an $\epsilon$-greedy policy where an action is performed randomly with probability $\epsilon$ and greedily otherwise: \(\pi(s)=\arg\max_{a\in\mathcal{A}}Q^{*}(s, a)\), where $Q^{*}(s, a)=\max_\pi Q^\pi(s,a)$.  The performance measure of policy \(\pi\) is defined as \(\mathcal{J} =\mathbb{E}_\pi[r(s, a)] \\= \mathbb{E}[\sum_{t=0}^{H-1}\gamma^tr(s_t, a_t)| \mathcal{P}, \pi]\). The objective is to determine a policy $\pi$ that maximizes \(\mathcal{J}\)
 through environment interaction.
 
\textbf{Double Deep Q-Learning. } Deep Reinforcement Learning with Double Q-Learning  \cite{van-hasselt-2010-DoubleQLearning}, combines the valuable properties of Double Q-Learning \cite{van-hasselt-2010-DoubleQLearning} and Deep Q-Learning \cite{mnih-DeepQ-Learning} in addressing the problem of Q values overestimation. DDQ-Learning approximates a parametric form of the state-action value function of policy $\pi$, $Q^\pi(s,a; \theta)$, with parameters $\theta$ often represented using neural networks. This is achieved through an iterative update procedure involving constant environment interaction. Provided an observation tuple $(s_t, a_t, r_{t+1}, s_{t+1})$ obtained at time step $t$, DDQ-Learning aims to approximate the true state-action value function by minimizing the following loss function for some tuple $i$:
\begin{equation}\label{eq:loss-ddql}
    \mathcal{L}_i(\theta_t)=[r_{t+1} + \gamma Q^\pi\big(s_{t+1}, \arg\max_{a^{'}}Q^\pi(s_{t+1}, a^{'}; \theta_t); \theta^{'}_t\big) - Q^\pi(s_t, a_t; \theta_t)]^2\,\,,
\end{equation}where $\theta_t$ corresponds to the parameters of $Q^\pi$. $\theta^{'}_t$ corresponds to the parameters the neural network in some previous time step. $\theta^{'}_t$ is a periodic copy of $\theta_t$. In practice, the loss function of Eq. (\ref{eq:loss-ddql}) is obtained as the expectation over a batch of observation tuples, sampled through prioritized experience replay \cite{schaul-2015-PER}.

\textbf{Prioritized experience replay. }During environment interaction the agent observes experience tuples comprised of $(s_t, a_t, r_{t+1}, s_{t+1})$ which are stored in a memory buffer $\mathcal{D}$. During DDQ-Learning a batch of such experiences is sampled using PER to update the network's parameters as proposed by \cite{schaul-2015-PER}. In the interest of space, we focus our discussion to \emph{proportional} prioritization as it is the method used in our work. Consider a priority value assigned to each tuple $i\in\mathcal{D}$, $p_i=\mathcal{L}_i + c$, where $c$ is a small constant. The probability of tuple $i$ to be sampled during training is $P(i) = \frac{p^\alpha_i}{\sum_{z\in\mathcal{D}} p^{\alpha}_z}$, where $\alpha$ is a problem dependent parameter determining the amount of prioritization. To compensate for the bias caused due to the frequency of sampling certain experience tuples, \cite{schaul-2015-PER} suggest the use of importance sampling weights of each tuple: $w_i=(\frac{1}{|\mathcal{D}|\cdot P(i)})^\beta$, where $|\mathcal{D}|$ is the size of the buffer and $\beta$ a  hyperparameter determining the extent to which an experience is weighted. In practice, $\beta$ is annealed from some initial $\beta_0$ to 1 and importance weights weigh gradient updates of the DDQ-Network parameters. 

\section{Control Problem Formulation}
    \textbf{Attractors. }Consider a PBN initialised at time step $t=0$ to some random state $s_0=\mathbf{e}$, that naturally evolves to some next state $s_{t+1}$ under no external perturbations, governed solely by its internal transition dynamics. Then, the PBN will eventually evolve to a set of states, that in Markov Chain theory are referred to as \emph{absorbing} \cite{schmulevich-pbn}. Absorbing states in PBNs correspond to \emph{attractors}. By the time a PBN enters an attractor, states outside it are no longer reachable without \textit{external} perturbations \cite{schmulevich-pbn}. However, such attractors in the context of GRNs reflect gene collective behaviour that exhibits biological properties that may be desirable or not. Hence, the ability to steer a PBN via \textit{external} perturbations towards an attractor with desirable properties is crucial in achieving a specific biological result, such as targeted therapeutics and cancer therapy.

\textbf{Control Framework. } Motivated by the biological properties exhibited by such attractors our objective is to determine the sequence of gene interventions that can drive the PBN from some state to a \textit{target} attractor.

\textbf{Definition 1. } \textit{Consider a PBN at time step $t$ with state $\mathbf{G}_t=\mathbf{e}$. Then we define intervention} $\textrm{I}(\mathbf{G}_t, u)$, $0\leq u \leq n$ \textit{at the state of the PBN, performed at time step $t$, as the process of flipping the binary value of the node $n_u$. $u=0$ denotes no intervention.}

At every time step only a single $\textrm{I}(\mathbf{G}_t, u)$ is allowed in order to ensure that our intervention method is the least intrusive to the network. Every intervention is followed by a natural PBN evolution step governed by the PBN's transition dynamics, which we do not consider when obtaining the control strategy. Then our objective is to determine the sequence $\mathbf{S}=\{\textrm{I}(\mathbf{G}_1, x), \textrm{I}(\mathbf{G}_2, y), \dots \textrm{I}(\mathbf{G}_{h\leq H}, z)\}$, where $x,y, ..., z\in\{k:0 \leq k \leq n\}$ in a restricted control horizon $H$, assuming that the MDP induced under the control framework is ergodic. Note that ergodicity is not a particularly restricting assumption and as shown experimentally in the next Section, successful control is achieved. 
\subsection{Connection to DDQ-Learning}
In order to construct a control strategy that successfully determines such sequence $\mathbf{S}$, we can translate our control problem to one of maximizing some performance metric $\mathcal{J}$ in the context of reinforcement learning that encapsulates our objective.

\textbf{Reward Scheme. }Correct reward assignment needs be made in order to approximate $Q^\pi(s,a; \theta)$ via DDQ-Network training, whose maximum expected reward would correspond to selecting interventions that drive the PBN to the target attractor - given a sequence of state observations. Consider $\mathcal{Y}$ to be the set of states of the target attractor, $s_t=\mathbf{e}$ and $a_t=u$. We define the following reward function:
\begin{equation}
r(s_t,a_t):=\begin{cases} 
      > 2 & \textrm{if } s_{t+1}\in\mathcal{Y}  \\
      -2 & \textrm{if } s_{t+1} \textrm{ is in a non-target attractor} \\
      -1 & \textrm{if } s_{t+1} \textrm{ is any other state} 
   \end{cases}
\end{equation}
While various reward assignments can prove useful, the proposed scheme as shown in the Experiments Section leads to successful control. The reason we assign $r(s_t, a_t)=-2$ to states in non-target attractors is to motivate our agent to avoid visiting states with undesirable biological properties, where it may get trapped. Initially during training successful control will rarely occur, hence, our motivation behind assigning $r(s_t, a_t)>2$ is to result in experience tuples with high priority values to be sampled more often through PER during DDQ-Learning. $r(s_t, a_t)=-1$ is assigned to all other states simply to motivate the agent to achieve control with the least possible interventions. In the case where only the target attractor is known the same reward scheme can be used, with $r(s_t, a_t)=-1$ assigned to all states, but the target attractor. Following the reward assignment defined in Eq. (2) maximizing the performance objective $\mathcal{J}$ corresponds to the process of finding a sequence of control interventions that drive the PBN to the target attractor $\mathcal{Y}$ with the least possible interventions.

\section{Experiments}
\subsection{Set Up}
To study the ability of a DDQ-Network with PER in learning to control PBNs under the framework introduced in Section 4, we implement the algorithm as proposed by \cite{ddql,schaul-2015-PER}. We evaluate its performance on 2 synthetically generated PBNs consisting of 10 and 20 nodes and a 7 node PBN inferred directly from gene expression data on a study of metastatic-melanoma \cite{bittner-2000-melanoma}. The hyperparameters used during training of the DDQ-Network are shown in Table 1, where $c$ refers to the time steps the values of $\theta_t$ are copied to $\theta_t^{'}$ and $r^\star(s,a) > 2$ the reward for successful control (Equation 2). Further, details on the structure of the PBNs and DDQ-Network with PER training parameters can be found in Appendix A.
\begin{table*}[tb] 
\small
\centering
\begin{tabularx}{\textwidth}{@{} lCCCCCC @{}}
      \Xhline{1\arrayrulewidth} \Tstrut
           \textbf{Environment} & Iterations & Horizon & $r^\star(s,a)$ & $|\mathcal{D}|$ & $\gamma$ & c\\
               \Xhline{1\arrayrulewidth} 

        PBN10 & 300,000 & 11 & 5& 1,024 & 0.95  & 500\Tstrut\\
        
        PBN20 & 700,000 & 100 & 20 & 50,000 & 0.90 &  5,000\Tstrut\\
        \Xhline{1\arrayrulewidth}

        & &&&&&\Tstrut \Bstrut\\
        \Xhline{1\arrayrulewidth}
       \Tstrut Melanoma & 150,000 & 7 & 5 & 1,024 & 0.99 &  500\Tstrut \Bstrut\\
        \Xhline{1\arrayrulewidth}
        & &&&&&\Tstrut \Bstrut\\

\end{tabularx}\caption{Hyperparameters for training a DDQ-Network to achieve successful control. }
\end{table*}

\textbf{PBN10. } In order to select our target attractor, we allow the PBN to naturally evolve to an attractor multiple times. We set our target to be the one occurring with the least frequency - hence, attempting to control for the hardest case. For PBN10 the selected target attractor has a probability of naturally occurring: $0.0097$ (for details, see Appendix A). Further, in order to set our allowed intervention horizon, we attempt to control the PBN10 towards the attractor via random interventions. We note that on average 1,387 interventions are required for successful control. Given that our objective is to minimize the number of interventions required to achieve control, we set our horizon to be approximately $0.8\%$ of the random interventions, namely 11 - allowing for a challenging control problem.

\textbf{PBN20. } We follow the same process as outlined in the PBN10 environment. Our selected target attractor naturally occurs with probability $0.00009$ and random interventions achieve control on average after 6,511 interventions. Hence, we set our maximum allowed horizon to $100$.

\textbf{Melanoma. } 
To further evaluate the control method on a real-world example, we construct a PBN from the gene expression data on the metastatic melanoma provided by Bittner {\it et al} \cite{bittner-2000-melanoma}. We infer a 7 node PBN with the exact 7 genes that appear in other studies, e.g., \cite{par-2006-pbn7,sirin-2013-BatchRL,kobayashi-2019-infer-pbn}, namely {\it pirin, WNT5A, S100P, RET1, MART1, HADHB, and STC2} (in that order). 
All data are initially discretized using \emph{median} quantization \cite{velarde2008}. Each node is assigned a set of boolean functions with varying gene predictors. Initially, all combinations of 3-gene predictors sets are evaluated on their ability to  predict the gene expression of a specific node. Their prediction accuracy is evaluated using the \emph{coefficient of determination} (COD) \cite{kim-2013-discovery}. The 10 predictors sets with the highest COD are selected as potential inputs for each node. A boolean function with its corresponding predictor set is randomly selected for each gene at every time step with a probability proportional to its COD.

Motivated by the argument that a deactivated WNT5A can reduce metastasis on the melanoma GRN \cite{datta-2003-control}, we set our target attractor to be an absorbing state with unexpressed WNT5A, namely:
$1\mathbf{0}01111$.
\begin{figure}[t!]
    \centering
    \includegraphics[width=\textwidth]{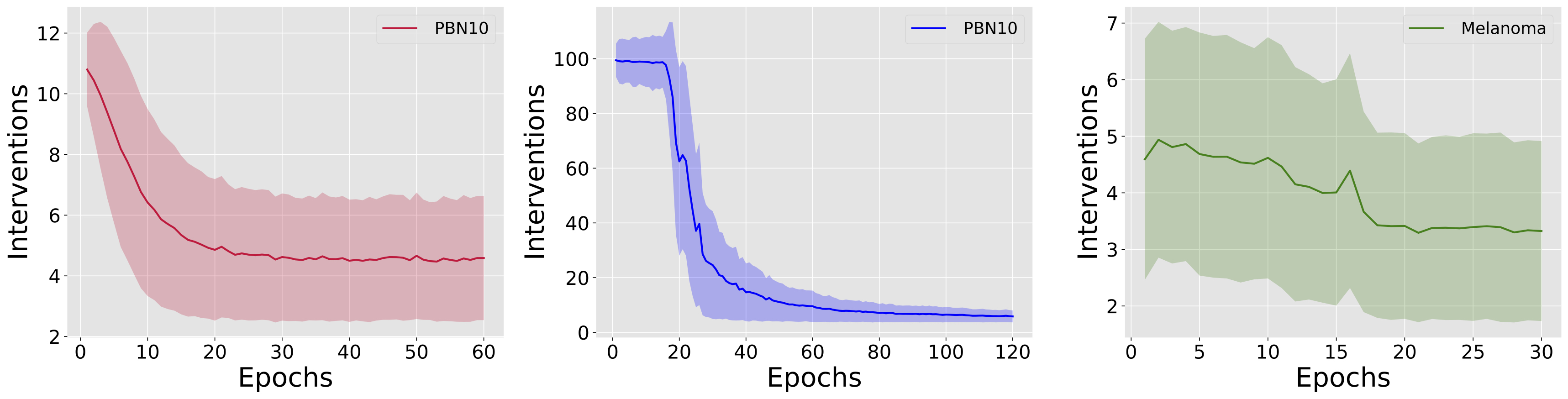}
    \caption{(1 Epoch = 5,000 Training Iterations) Mean and standard deviation of the number of interventions performed during training obtained for every epoch.}
\end{figure}{}

\subsection{Results}
Figure 1 demonstrates the number of interventions required during training for the DDQ-Network to achieve control or stop the attempt if the maximum number of interventions is reached. The number of interventions begins to decrease sharply after the first few training iterations. As shown, the DDQ-Network determines control strategies that drive the PBNs to the target attractors with \emph{significantly} less number of interventions compared to the specified horizon of allowed interventions. Note however that the results shown in Figure 1 correspond to the interventions made by an $\epsilon$-greedy policy. Hence, the results also account for random interventions taken during training that favour exploration and can hinder performance especially near the end of training. 

\textbf{Evaluation. }To properly test the performance of the obtained control strategies we randomly initialize each PBN from some random state 10,000 times and attempt to control the PBN towards its corresponding target attractor. During testing we use a \emph{greedy} policy. At each time step we perform the intervention yielding the maximum expected reward according to the state-action values approximated by the DDQ-Network. 

We note that for the PBN10 environment the DDQ-Network obtains successful control with a $99.8\%$ success rate when allowed a maximum of 11 interventions. During testing $100\%$ successful control rate was achieved when 14 interventions were allowed.

For the PBN with 20 nodes the same evaluation process is followed. We note successful control with $100\%$ rate when allowed 100 interventions. Interestingly, approximately $99\%$ successful control occurs with up to 15 interventions, as also shown in Figure 1.

Finally, for the real-world case of metastatic-melanoma we note that the DDQ-Network can drive the PBN to the attractor with unexpressed WNT5A with a success rate of $99.52\%$ when allowed 7 interventions and can reach up to $99.9\%$ when 10 interventions are permitted.

\textbf{Comparison. } Making a direct comparison with previous learning
approaches such as those mentioned in Section 2 is not possible
due to the notion of control we adopt that draws inspiration from complex systems \cite{liu-2011-controllability} that has only been applied to BNs \cite{karlsen-2018-evolution,papagiannis-2019-drl-rbn} and not PBNs.  Other similar works discussed show results of
regulating one gene or attempting control with a pre-specified number of control inputs. However, it is interesting to note that compared
to the approach of XCS \cite{karlsen-2018-evolution}, due to scalability issues faced by
Learning Classifier Systems, XCS was limited to RBN of size
$n$ = 9, modelling the cell cycle of fission yeast \cite{karlsen-2019-yeast}. Instead we worked with significantly more complex PBNs up to $n$ = 20, hence increasing the state space by a factor of 32,768. 

Further, the PBNs, compared to RBNs, introduced a factor of high stochasticity that can lead to significantly more challenging control problems. While \cite{papagiannis-2019-drl-rbn} shows successful application of DDQ-Learning with PER to networks of 25 nodes, the experiment setting is simpler as it explicitly focuses on RBNs.  Further, in relation to the Batch Reinforcement Learning method in \cite{sirin-2013-BatchRL} while the authors show successful results on a melanoma GRN with $n=28$ their target states comprise half of the state space, naturally occurring with a probability of $0.54$ compared to our experiments where the target states comprise one state naturally occurring with probabilities as low as $0.00009$. Therefore, while a direct comparison is not feasible due to the difference in the control framework adopted, it can be seen that the DDQ-Learning with PER method is very robust and can handle complex environments.

\textbf{Discussion. }As demonstrated through the experiments performed on the three PBNs, DDQ-Learning with PER  successfully constructs control strategies to drive the networks to their target attractor. The strategies are learned through direct environment interactions, without the need to infer or utilize the dynamics of the networks. This is crucial, as model-free methods can be useful when dealing with networks whose dynamics are unknown or difficult to infer and only a target state is available. Further, we note that the reward scheme introduced in Eq. (2) motivates the agent during training to determine sequences of control interventions that are significantly less than the initially allowed control horizon. Also, after evaluating the control method we observe that the agent is very robust to the high stochasticity of the constructed PBNs as it achieves control with high success rate after initializing it from multiple random states.

\section{Conclusion}
In this work we have studied the ability of the Double Deep Q-Learning with Prioritised Experience Replay method \cite{ddql,schaul-2015-PER} in learning how to control Probabilistic Boolean Networks. The applied method is model free and develops control strategies agnostic to the structure and underlying dynamics of the networks, directly from state observations. We demonstrate through extensive experiments on two highly stochastic PBNs and a PBN inferred from real-world gene data that the method can efficiently learn how to steer the PBNs to the least naturally occurring attractors with high success rate. The method we applied can be suitable for obtaining control strategies in problems were the only information available is a target state exhibiting favourable biological properties, bypassing the extensive step of utilizing and inferring the PBNs' transition dynamics.

Learning control strategies in a model-free manner can be further studied from several directions. Currently, a limitation of the method is that it learns how to control in a sample inefficient manner, requiring multiple environment interactions in order to build optimal policies. An interesting extension would be to determine ways to utilize the PBN's transition patterns observed during training to improve the learning efficiency. Further, during testing, we note that the DDQ-Network achieves control by perturbing the value of a specific subset of the available nodes. Further study could provide insights into alternative ways of identifying {\it control nodes} \cite{liu-2011-controllability,gao-2014-target-control,moschoyiannis-2016-control,choo-2018-control-nodes} in PBNs.

\bibliographystyle{splncs03}
\bibliography{drl-pbn}
\newpage
\appendix
\section{Experiments: Supplementary details}
In this section we describe the exact details of the two synthetic PBNs used in our experiments. The dynamics are shown to make our results reproducible, however, they are not used as part of the learning algorithm. PBN10 is shown in Table 2 and PBN20 in Table 3. Initially, we use the respective transition probabilities in order to build state transition graphs. The state transition graphs are then used to determine attractors. Finally, we randomly evolve each network to determine which attractors naturally occur with the least frequency. We then set those attractors as our targets.

\begin{table*} 
\small
\centering
\begin{tabularx}{\textwidth}{@{} lCCC|CCCC @{}}
      \Xhline{1\arrayrulewidth} \Tstrut
           \textbf{Function Set} & OR & AND & XOR & & OR & AND & XOR \TBstrut\\
               \Xhline{1\arrayrulewidth} 

        $\mathbf{F}_1$ & 1.00 & - & - &  $\mathbf{F}_{6}$ & 0.82 & 0.15 & 0.03      \Tstrut\\
        $\mathbf{F}_2$ & 0.50 & 0.25 & 0.25    &   $\mathbf{F}_{7}$ & 0.48 & 0.52 & -           \Tstrut\\
        $\mathbf{F}_3$ & 0.71 & 0.29 & -       &  $\mathbf{F}_{8}$ & 0.28 & 0.45 & 0.27            \Tstrut\\
        $\mathbf{F}_4$ & 0.52 & 0.48 & -    &  $\mathbf{F}_{9}$ & 1.00 & - & -                 \Tstrut\\
        $\mathbf{F}_{5}$& 0.36 &0.05 & 0.59   &  $\mathbf{F}_{10}$ & 0.99 & 0.01 & -         \Tstrut\\    

        \textbf{Probabilities} & $p^i_1$ & $p^i_2$ & $p^i_3$ &&  $p^i_1$ & $p^i_2$ & $p^i_3$\Tstrut\\
         & & & \Tstrut\\

\end{tabularx}\caption{State transition probabilities for the PBN10 environment.}
\end{table*}

\vspace{-30pt}
\begin{table*} 
\small
\centering
\begin{tabularx}{\textwidth}{@{} lCCC|CCCC @{}}
      \Xhline{1\arrayrulewidth} \Tstrut
           \textbf{Function Set} & OR & AND & XOR & & OR & AND & XOR \TBstrut\\
               \Xhline{1\arrayrulewidth} 

        $\mathbf{F}_1$ & 0.39 & 0.04 & 0.57 &  $\mathbf{F}_{11}$ & - & 1.00 & -      \Tstrut\\
        $\mathbf{F}_2$ & 0.70 & - & 0.30    &   $\mathbf{F}_{12}$ & - & 1.00 & -           \Tstrut\\
        $\mathbf{F}_3$ & 1.00 & - & -       &  $\mathbf{F}_{13}$ & 1.00 & - & -            \Tstrut\\
        $\mathbf{F}_4$ & 0.18 & 0.82 & -    &  $\mathbf{F}_{14}$ & 0.01 & 0.98 & 0.01                 \Tstrut\\
        $\mathbf{F}_{5}$& - & 0.11 & 0.89   &  $\mathbf{F}_{15}$ & - & - & 1.00                 \Tstrut\\
        $\mathbf{F}_6$ & 1 & - & -          & $\mathbf{F}_{16}$ & - & 1.00 & -            \Tstrut\\
        $\mathbf{F}_7$ & 1 & - & -          &  $\mathbf{F}_{17}$ & 1.00 & - & -            \Tstrut\\
        $\mathbf{F}_8$ & - & 0.44 & 0.56    & $\mathbf{F}_{18}$ & - & 1.00 & -                 \Tstrut\\
        $\mathbf{F}_9$ & - & - & 1.00       & $\mathbf{F}_{19}$ & - & - & 1.00                 \Tstrut\\
        $\mathbf{F}_{10}$ & 0.82 & 0.09 & 0.09 & $\mathbf{F}_{20}$ & 1.00 & - & -                 \Tstrut\\
        \textbf{Probabilities} & $p^i_1$ & $p^i_2$ & $p^i_3$ &&  $p^i_1$ & $p^i_2$ & $p^i_3$\Tstrut\\
         & & & \Tstrut\\

\end{tabularx}\caption{State transition probabilities for the PBN20 environment.}
\end{table*}
\vspace{-20pt}
\textbf{Training. } We constructed a deep neural network with an input layer of size  $n$, two hidden layers each of 100 rectifier units and a linear output unit of size $n+1$ corresponding to the expected $Q^\pi$ values of the possible network interventions (including no intervention). During training we use an \(\epsilon\)-greedy policy that randomly performs an action with probability $\epsilon$, otherwise selects an intervention greedily by taking the maximum over $Q^\pi$. Further, we set $\beta_0=0.4,\,\, \alpha=0.6$ for PER.

\end{document}